# A Constructive Algorithm for Feedforward Neural Networks for Medical Diagnostic Reasoning

Abu Bakar Siddiquee, Md. Ehsanul Hoque Mazumder and S. M. Kamruzzaman

Department of Computer Science and Engineering

International Islamic University Chittagong, Bangladesh.

Email: maksud_cse@yahoo.com

*Abstract -* **This research is to search for alternatives to the resolution of complex medical diagnosis where human knowledge should be apprehended in a general fashion. Successful application examples show that human diagnostic capabilities are significantly worse than the neural diagnostic system. Our research describes a constructive neural network algorithm with backpropagation; offer an approach for the incremental construction of near-minimal neural network architectures for pattern classification. The algorithm starts with minimal number of hidden units in the single hidden layer; additional units are added to the hidden layer one at a time to improve the accuracy of the network and to get an optimal size of a neural network. Our algorithm was tested on several benchmarking classification problems including Cancer1, Heart, and Diabetes with good generalization ability.**

## I. INTRODUCTION

The main characteristic of neural networks (NN) is their ability to generalize information, as well as their tolerance to noise. Therefore, one of the computer science areas that use them the most is Pattern Recognition. The research line of Neural Networks applied to Pattern Recognition has as its main objectives:
o The application of Neural Networks to specific problems of pattern recognition.
o The evaluation of efficiency metrics and reliability of the solutions proposed.
o The formation of human resources in the area.

Backpropagation algorithm is the most widely used learning algorithm to train multiplayer feedforward network and applied for applications like character recognition, image processing, pattern classification etc. Before train an artificial neural network (ANN), the network must be built. That is, the nodes in the input layer, output layer and the hidden layer must be defined. The ANNs used for the same problem may differ from each other by their length of hidden layer as the length of hidden layer has to be pre-determined in the traditional backpropagation algorithm. In recent years, many researches have been done on algorithms that dynamically build neural networks for solving pattern classification problems. These algorithms include the dynamic node creation, the cascade correlation algorithm, the self-organizing neural network, and the upstart algorithm. In this research we also proposed an algorithm, which can add nodes in the single hidden layer during the training period and can build an ANN with its minimal size, which can classify patterns with acceptable efficiency.

## II. NEW CONSTRUCTIVE ALGORITHM

One of the problems with the traditional backpropagation algorithm is the predetermination of the number of neurons in the hidden layer within a network. To overcome this problem the *construction algorithm* for feedforward networks may be used, which constructs the network during training. Thus we can have an optimal number of neurons in the hidden layer to attain a satisfactory level of efficiency for a particular problem. Besides applying the *early stopping method* of training using *cross-validation* we can also train the network in a relatively short estimation period (training period). In the construction algorithm proposed by Rudy Setiono and Huan Liu they have defined the stopping condition of the training by classifying all the input patterns. It means that while the efficiency is 100%, the training will stop. But in most cases with the benchmarking classification problems 100% efficiency may not be achieved. This is why we used a new algorithm for pattern classification that defines the stopping condition by the acceptance of efficiency level. Another consideration we have made that the desired or acceptable efficiency on the test sets may not be achieved even though the *mean square error* on training set is minimum. These considerations encouraged us to propose an algorithm that will combine the learning rule of backpropagation algorithm to update weights of the network and the construction algorithm to construct the network dynamically and also consider the efficiency factor as a determinant of the training process.

The following steps are followed to build and train a network [5];
1. Create an initial neural network with number of hidden unit h = 1. Set all the initial weights of the network randomly within a certain range.
2. Train the network on training set by using a training algorithm for a certain number of epochs that minimizes the error function.
3. If the error function $\xi_{av}$ on validation set is acceptable and, at this position, the network classifies desired number of patterns on test set that leads the efficiency *E* to be acceptable then *stop*.
4. Add one hidden unit to hidden layer. Randomly initialize the weights of the arcs connecting this new hidden unit with input nodes and output unit(s). Set h = h + 1 and go to *step 2*.

For back propagation algorithm the weight adjustment is:
For the output-layer weights:

$$w^o_{kj_{new}} = w^o_{kj_{old}} + \eta \delta^o_{pk} i_{pj} \quad \text{---(1)}$$

Where, $\delta^o_{pk} = \delta_{pk} f^{o'}_k (net^o_{pk})$

For the hidden-layer weights:

$$w^h_{ji_{new}} = w^h_{ji_{old}} + \eta \delta^h_{pj} x_{pi} \quad \text{---(2)}$$

Where, $\delta^h_{pj} = f^{h'}_j (net^h_{pj}) \sum_k \delta^o_{pk} w^o_{kj}$

Where, *k* indicates the *k*th output unit, *j* indicates the *j*th hidden unit; *i* indicates the *i*th input node, *p* is the input vector, *η* is the learning rate, *δ* is the error term, $x_{pi}$ is the input value to the *i*, $f^{o'}_k (net^o_{pk})$ is the output function of the *k*th output unit, $f^{h'}_j (net^h_{pj})$ is the output function of *j* connected to *k*

The error function is usually defined as the mean-squared-errors

$$e_k = d_k(n) - y_k(n) \quad \text{---(3)}$$

$$\xi(n) = \frac{1}{2} \sum_{k \in C} e_k^2(n) \quad \text{---(4)}$$

$$\xi_{av} = \frac{1}{N} \sum_{n=1}^{N} \xi(n) \quad \text{---(5)}$$

Where, k denotes kth output unit, n denotes the nth iteration, C is the number of output units, N is the total number of patterns, $d_k$ denotes the desired output from k, $y_k$ denotes the actual output of neuron k, $e_k$ denotes the error term for kth output unit.

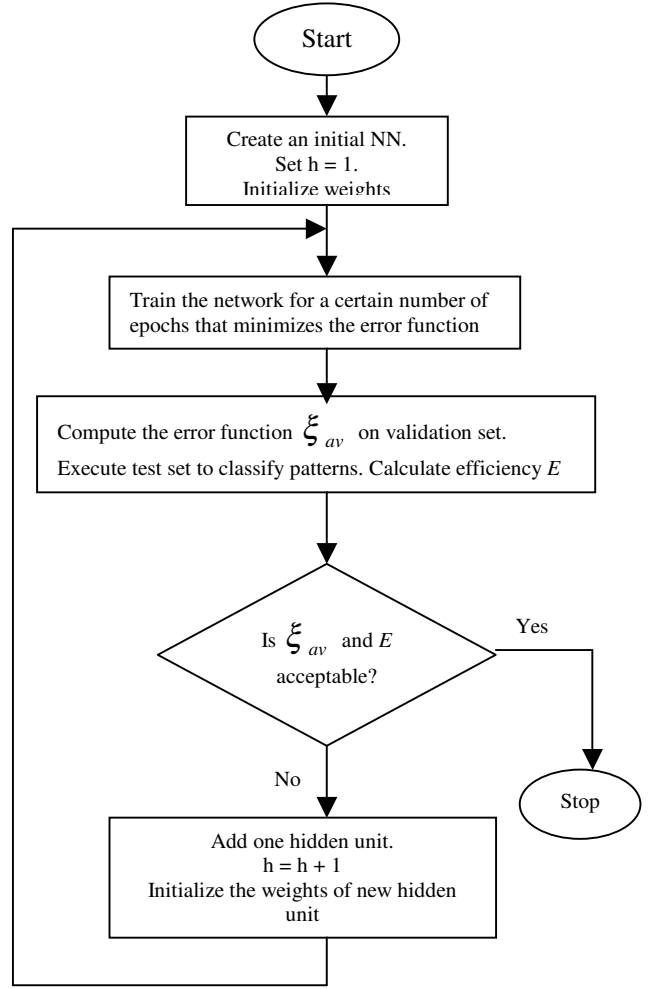

**Figure 1** Flow chart of the Proposed Algorithm

## III. EXPERIMENTAL RESULTS

We tested our algorithm by three benchmark classification problems Cance1, Heart, Diabetes. The goal is to show that by using minimal size of a neural network we can classify patterns in medical diagnostic data sets with an acceptable efficiency and can also minimize the training period. Table shows the results with different number of nodes in the hidden layer. The bolded rows show the optimal size of the hidden layer with best result.

TABLE 1
CHARACTERISTICS OF DATA SETS OF CANCER1, HEART, DIABETES TO SHOW THE NUMBER OF EXAMPLES IN TRAINING, VALIDATION AND TEST SET

|  | INPUT ATTRIB | OUT UNITS | OUT CLASSES | TRAIN EXAMP. | VALID. EXAMP. | TEST EXAMP. | TOTAL EXAMP |
|---|---|---|---|---|---|---|---|
| **Cancer1** | 9 | 2 | 2 | 350 | 175 | 174 | 699 |
| **Heart** | 13 | 1 | 2 | 152 | 76 | 75 | 303 |
| **Diabetes** | 8 | 2 | 2 | 384 | 192 | 192 | 768 |

TABLE 2
EXPERIMENTAL RESULTS SHOWS THE BEST FITTED NETWORK BY THE BOLDED ROWS FOR EACH DATA SET

|  | NO OF HU | EPOCH | Training set ||| Valid set ||| Test set ||| OVERALL EFFICIENCY |
|---|---|---|---|---|---|---|---|---|---|---|---|---|
|  |  |  | CLASSIFIED | EFFICIENCY | MEAN SQUARE ERROR | CLASSIFIED | EFFICIENCY | MEAN SQUARE ERROR | CLASSIFIED | EFFICIENCY |  |  |
| **Cancer1** | 1 | 100 | 338 | 96.57 | 0.0156 | 169 | 96.00% | 0.0171 | 172 | 98.00% | 97.13877 |
|  | **2** | **450** | **339** | **96.86** | **0.0138** | **167** | **95.00%** | **0.0229** | **173** | **99.00%** | **97.13877** |
|  | 3 | 550 | 339 | 96.86 | 0.0136 | 167 | 95.00% | 0.0229 | 172 | 98.00% | 96.99571 |
|  | 4 | 1350 | 346 | 98.86 | 0.008 | 169 | 96.00% | 0.0171 | 168 | 96.00% | 97.71102 |
| **Heart** | 1 | 500 | 143 | 93.46 | 0.0395 | 64 | 85.00% | 0.0855 | 60 | 80.00% | 88.11881 |
|  | 2 | 600 | 144 | 94.12 | 0.0395 | 63 | 84.00% | 0.0855 | 60 | 80.00% | 88.11881 |
|  | 3 | 700 | 139 | 90.85 | 0.0428 | 63 | 84.00% | 0.0987 | 60 | 80.00% | 86.46865 |
|  | **4** | **800** | **142** | **92.81** | **0.0428** | **62** | **82.00%** | **0.1118** | **61** | **81.00%** | **87.45875** |
| **Diabetes** | 1 | 200 | 300 | 78.12 | 0.2083 | 141 | 73.00% | 0.2656 | 132 | 68.00% | 74.60938 |
|  | 2 | 380 | 312 | 81.25 | 0.1875 | 147 | 76.00% | 0.2344 | 135 | 70.00% | 77.34375 |
|  | 3 | 480 | 311 | 80.99 | 0.1875 | 147 | 76.00% | 0.2344 | 135 | 70.00% | 77.21354 |
|  | **4** | **1580** | **324** | **84.38** | **0.151** | **148** | **77.00%** | **0.2292** | **143** | **74.00%** | **80.07813** |
|  | 5 | 1900 | 327 | 85.16 | 0.151 | 147 | 76.00% | 0.2396 | 130 | 67.00% | 74.60938 |

## IV. COMPARATIVE STUDY

As concentrating on the optimal size of an artificial neural network, we have ignored all the additional features (like, error smoothing functions, momentum constant, weight freezing technique etc.) that are used to increase the performance of a backpropagation algorithm, to examine, the level of efficiency of a network with optimal size. Comparing with the previous works on the benchmarking problems we have used and their results, the classification efficiency of our proposed algorithm is quite acceptable. Table 3 shows the comparisons.

## V. CONCLUSION

We have examined the experimental results by using different learning constant. Hence we can say that for these data sets used, better results are achieved when the value of learning constant ($\eta$) is 0.7 or 0.8. The learning constant used beyond this range did not cause better efficiency. In our experiment we did not use the momentum constant ($\alpha$). As we were focusing at the optimal size of a network that performs the pattern classification with acceptable efficiency, we have ignored all the factors that improve the performance of backpropagation algorithm. The reason is to examine how optimal the size of a network can be with the use of backpropagation. The proposed algorithm has worked with a better performance. It is that the overall performance is quite acceptable. Many researchers have proposed different techniques to improve backpropagation (smoothing functions of error term, weight freezing during training, combination of Kohonen layer with backpropagation). Using any one or more of these techniques can improve the performance of the proposed algorithm. By using cross validation the early stopping point was also found in our experiment. This also reduced the period of training.

TABLE 3
COMPARISON WITH OTHER WORKS BASED ON EFFICIENCY

| Researcher(s) | Research work | Efficiency | Average Efficiency of Proposed algorithm on test set |
|---|---|---|---|
| **Cancer1** | | | |
| Rudy Setiono (1996) | Extracting rules from pruned neural networks for breast cancer diagnosis | 93.69% | 97.5% |
| **Heart** | | | |
| R.Detrano, A. Janosi, W. Steinbrunn, M. Pfisterer, J. Schmid, S. Sandhu, K. Guppy, S. Lee, V. Froelicher (1989). | International application of a new probability algorithm for the diagnosis of coronary artery disease. | 77% | 80% |
| David W. Aha Dennis Kibler | Instance-based prediction of heart disease presence with the Cleveland database | 77% | 80% |
| J.H. Gennari P. Langley (1989) | Incremental concept formation | 78.9% | 80% |
| **Diabetes** | | | |
| J.W. Smith, J. E. Everhart, W.C. Dickson, W. C. Knowler, R. S. Johannes, (1988). | ADAP learning algorithm to forecast the onset of diabetes mellitus | 76% | 71% |